\documentclass[nohyperref]{article}
\usepackage{microtype}
\usepackage{graphicx}
\usepackage{subfigure}
\usepackage{booktabs} 

\usepackage{hyperref}
\usepackage{algorithm}
\usepackage{algorithmic}
\usepackage{bm}

\usepackage[accepted]{icml2023}

\usepackage{amsmath}
\usepackage{amssymb}
\usepackage{mathtools}
\usepackage{amsthm}
\usepackage{stfloats}

\usepackage[capitalize,noabbrev]{cleveref}

\theoremstyle{plain}

\theoremstyle{definition}

\theoremstyle{remark}

\usepackage[textsize=tiny]{todonotes}

\icmltitlerunning{}

\begin{document}
\twocolumn[
\icmltitle{Intrinsic Bayesian Optimisation on Complex Constrained Domain}




\begin{icmlauthorlist}
\icmlauthor{Yuan Liu}{}
\icmlauthor{Mu Niu}{}
\icmlauthor{Claire Miller}{}
\end{icmlauthorlist}


\icmlcorrespondingauthor{Mu Niu}{Mu.Niu@glasgow.ac.uk}

\icmlkeywords{Bayesian Optimisation, Intrinsic Gaussian Processes, Riemannian geometry, Manifolds, Brownian Motion, Heat kernel.}

\vskip 0.3in
]



\printAffiliationsAndNotice{\icmlEqualContribution} 

\begin{abstract}
Motivated by the success of Bayesian optimisation algorithms in the Euclidean space, we propose a novel approach to construct Intrinsic Bayesian optimisation (In-BO) on manifolds with a primary focus on complex constrained domains or irregular-shaped spaces  arising as submanifolds of $\mathbb{R}^2$, $\mathbb{R}^3$ and beyond. Data may be collected in a spatial domain but restricted to a complex or intricately structured region corresponding to a geographic feature, such as lakes. Traditional Bayesian Optimisation (Tra-BO) defined with a Radial basis function (RBF) kernel cannot accommodate these complex constrained conditions. The In-BO uses the Sparse Intrinsic Gaussian Processes (SIn-GP) surrogate model to take into account the geometric structure of the manifold. SIn-GPs are constructed using the heat kernel of the manifold which is  estimated as the transition density of the Brownian Motion on manifolds. The efficiency of In-BO is demonstrated through simulation studies on a U-shaped domain, a Bitten torus, and a real dataset from the Aral sea. Its performance is compared to that of traditional BO, which is defined in Euclidean space.
\end{abstract}

\section{Introduction}
\label{submission}
Optimization problems arise in all quantitative disciplines from statistics and computer science to engineering and economics. In recent years, the optimization objective is no longer limited to Euclidean space with simple spatial structure. 
Optimization problems on manifolds have gradually become a research hotspot. We suppose that observations can only be collected from grid points on the manifold. The problem can be posed from the perspective of optimization as trying to find the optimal point within these grid points so that the corresponding implicit objective function reaches its global maximum (or minimum). Let $M$ be a $d$ dimensional complete Riemannian manifold. It is also the submanifold of a higher dimensional Euclidean space $R^p$, $d\leq p$. The problem can be defined as:
\\
\rule[0pt]{8.3cm}{0.05em}
\\
\\
\textbf{Define:}  $f(s)$ is the objective function defined on $M$, which does not have an explicit expression. Solve:
\begin{equation}
s_{M}=\operatorname{argmax}_{s \in S} f(s),
\end{equation}
where S is a set of grid points defined on $M$.
\\   
\rule[0pt]{8.3cm}{0.05em}

To better illustrate the problem, imagine we need to find the location with the highest chlorophyll levels in the Aral Sea \cite{Aralsea01} in Figure 1. The distribution of the chlorophyll level in the constrained domain is unknown and can be treated as a blackbox function. The geometry of the constrained domain is also different from the Euclidean space $R^2$. Two locations that have a close Euclidean distance on a map may be intrinsically far apart if they are separated by a land barrier.

Bayesian optimization (BO) is an effective solution to solve optimisation problems in Euclidean space when the objective function is unknown \cite{Shahriari2016}. It outperforms other leading-edge global optimization algorithms on a wide range of challenging optimization benchmark functions \cite{jones2001taxonomy}. The Traditional Bayesian optimization (Tra-BO) framework consists of two core components, the probabilistic surrogate model and the acquisition function. The probabilistic surrogate model is used to approximate the unknown function to be optimized. Gaussian Processes (GP), known for its high flexibility, has been commonly utilized as a surrogate model in BO due to its ability to theoretically model a wide range of objective functions. The acquisition function is used to decide where to sample next by taking into account both the current best estimate of the function and the uncertainty in that estimate, balancing the trade-off between exploration and exploitation. There are many different types of acquisition functions that have been proposed. PI quantifies the probability that an observation of $x$ will improve the value of the current optimal objective function \cite{kushner1964new}. Močkus et al. \cite{mockus1978application} proposed a new improvement-based strategy: Expected improvement (EI). Srinivas et al. \cite{srinivas2009gaussian} proposed a placement strategy for Gaussian processes: GP-UCB, where UCB means Upper Confidence Bound. We focus on the Probability of Improvement (PI) in this work, which selects the next point to sample based on the probability that the function value at that point will be greater than the current best observed value \cite{kushner1964new}. 
\begin{figure}[ht]
\vskip 0.2in
\begin{center}
\centerline{\includegraphics[width=2.5in]{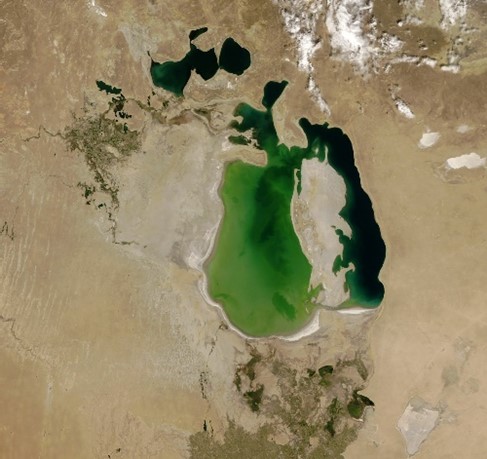}}
\caption{Satellite imagery of the Aral Sea, an endorheic basin (saltwater lake) in Central Asia.\cite{Aralsea01}}
\label{icml-historical}
\end{center}
\vskip -0.2in
\end{figure}

Our work concentrates on the optimisation problem on manifolds with geometric constraints. A manifold is a topological space that locally resembles Euclidean space near each point. The geometry of a manifold is in general different from the usual Euclidean geometry. For example, in the Aral sea shown in Figure 1, data are collected in this spatial domain, restricted to the complex or intricately structured region. It is important to take into account the intrinsic geometry of the sea and its complex and irregular boundary. Traditional smoothing or modelling methods that do not respect the intrinsic geometry of the space, and especially the boundary constraints, may produce poor predictions \cite{niu2019intrinsic}. Tra-BO based on Euclidean Gaussian Processes defined with an RBF kernel, cannot recognize the fact that pairs of locations having close Euclidean distance may be intrinsically far apart if separated by boundaries. We propose a general methodology, named Intrinsic Bayesian Optimisation (In-BO), to solve optimisation problems in complex spatial subregions of $\mathbb{R}^{2}$ and $\mathbb{R}^{3}$.

In contrast to Tra-BO,  the proposed In-BO on manifold approach employs an Intrinsic Sparse Gaussian Process (SIn-GPs) on manifolds as the probabilistic surrogate model. In-BO will be guided by the posterior predictive distribution of the SIn-GPs to explore the manifold. It can accommodate the interior structure of the manifold and respect the boundaries. In-BO can enhance optimization efficiency and precision by utilizing the combination of SIn-GP on manifolds and the probabilistic improvement acquisition function. We organize our works as following. In Sec.2, we introduce the concepts of Riemannian geometry. In Sec.3, we present a concrete illustration of our In-BO algorithm, including updating with sparse In-GP on manifolds and estimating the heat kernel as the transition density of Brownian Motion(BM) on manifolds. In Sec.4, the In-BO algorithm is employed to tackle optimization challenges on the U-shape domain, the Bitten Torus, and a real-world dataset of the Aral Sea. The performance of the proposed In-BO method is evaluated against the Tra-BO.


\section{Riemannian Geometry}

A manifold, is often defined as a topological space where each points in this space belongs to a neighbourhood that is homeomorphic to an open subset of n-dimensional Euclidean space \cite{chen1999lectures}. Let $M$ be a d-dimensional complete and orientable Riemannian manifold, $\partial M$ is $M$'s boundary which is continuous and $C^{1}$ almost everywhere, and $\delta $ is the Dirac delta function. 

\textbf{Heat Kernel}: The heat kernel $K_{h e a t}\left(s_{0}, s, t\right)$ describes the process of heat diffusion on the manifold $M$,  given a heat source $s_{0}$ and time $t$. This heat kernel should satisfy the heat equation:
\begin{equation}
\label{eqn:heat}
\begin{aligned}
\frac{\partial}{\partial t} K_{h e a t}\left(s_{0}, s, t\right)=\frac{1}{2} \Delta_{s} K_{h e a t}\left(s_{0}, s, t\right),\\
\lim _{t \rightarrow 0} K_{\text {heat }}\left(s_{0}, s, 0\right)=\delta\left(s_{0}, s\right), \quad s_{0}, s \in M
\end{aligned}
\end{equation}
where $\Delta s$ is the Laplacian-Beltrami operator on $M$. The initial condition is valid in a distributional sense \cite{berline2003heat}. $K_{h e a t}\left(s_{0}, s, t\right)$ is a smooth function on $M \times M \times R^{+}$ which satisfies the heat equation. It is symmetric and is a positive semidefinite kernel on $M$. We define the Neumann boundary condition along $\partial M$ to setup no heat transfers across the boundary $\partial M$:
\begin{equation}
\frac{\partial K}{\partial \mathbf{n}}=0 \quad \text { along } \partial M \text {, }
\end{equation}
where $n$ denotes a normal vector of $\partial M$. The heat kernel is analytically intractable to directly evaluate for general Riemannian manifolds. The closed-form expression is only available for the special case like Euclidean space. If $M$ is $R^{d}$. The heat kernel of $M$ can be expressed as:
\begin{equation}
K_{\text {heat }}\left(\mathbf{s}_0, \mathbf{s}, t\right)=\frac{1}{(2 \pi t)^{d / 2}} \exp \left\{-\frac{\left\|\mathbf{s}_0-\mathbf{s}\right\|^2}{2 t}\right\}, \mathbf{s} \in \mathbb{R}^d,
\end{equation}
which is the scaled version of an RBF kernel with different parametrisations:
\begin{equation}
K_{R B F}\left(\mathbf{x}_{0}, \mathbf{x}, l\right)=\sigma_{r}^{2} \exp \left\{-\frac{\left\|\mathbf{x}_{0}-\mathbf{x}\right\|^{2}}{2 l^{2}}\right\}, \mathbf{x} \in \mathbb{R}^{d}.
\end{equation}
 
 The time parameter $t$ of $K_{\text {heat }}$ has the same effect as that of the length scale parameter $l$ of $K_{R B F}$, controlling the rate of decay of the heat kernel. Let $K_{\text {heat }}^{t}(s_0, s)=K^{\text {heat }}(s_0, s, t)$. The explicit expression for the heat kernel $K_{\text {heat }}^{t}$ on general Riemannian manifolds does not exist. To circumvent solving the heat equation on manifolds, heat kernels can also be interpreted as transition densities of Brownian Motion (BM) on $M$.
 

 \textbf{Brownian Motion on $M$}. The Riemannian manifold $M$ is equipped with a metric tensor $g$. Let $\phi: \mathbb{R}^{d} \rightarrow M$ be a local parametrisation of $M$. The metric tensor can be represented as a symmetric, positive definite matrix-valued function, which defines a smoothly varying inner product in the tangent space of $M$. It can be described in local coordinates as 
 \begin{equation}
g_{ij} = \frac{\partial{\phi(x)}}{\partial x_i} \frac{\partial{\phi(x)}}{\partial x_j}.
 \end{equation}

\begin{figure*}[ht]
	\centering
     \centering
\includegraphics[height=4.5cm]{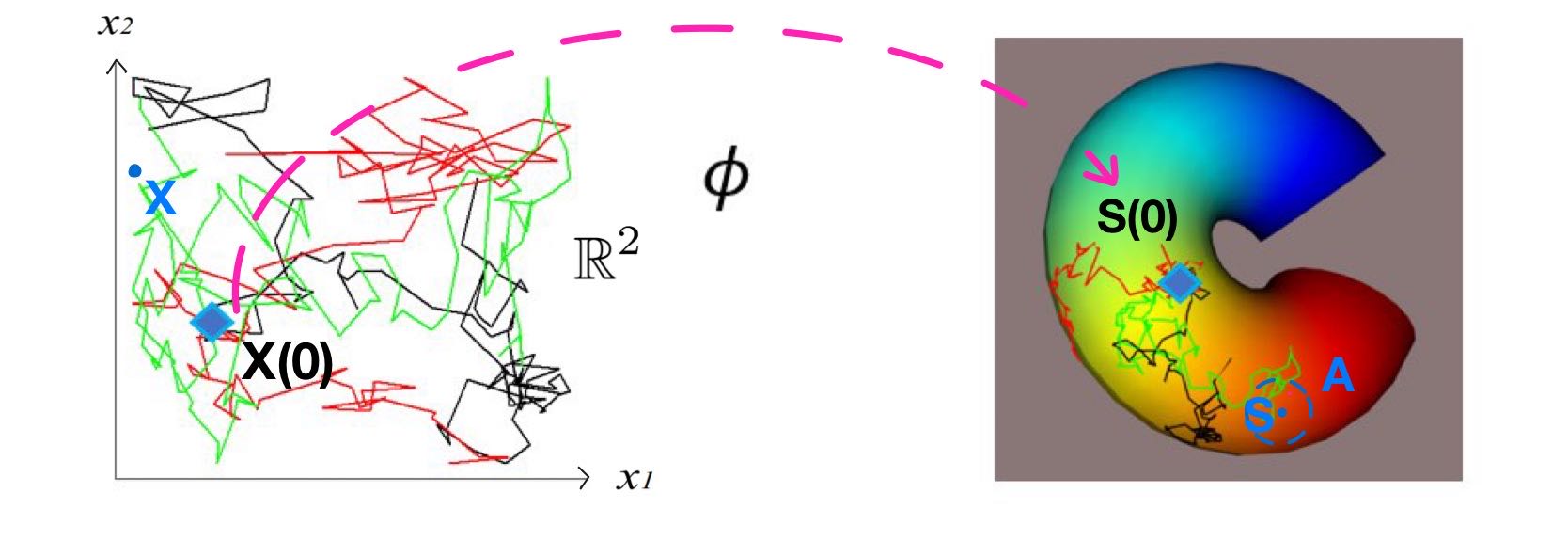}
\caption{ \label{fig:bm}BM on Bitten Torus and its equivalent stochastic process in $\mathbb{R}^{2} $: Three BM sample paths from same initial point $S_{0}$, shown in different colours; only the green sample path reaches $A$ at time $t$. $\phi: \mathbb{R}^{2} \rightarrow M$ is a local parametrisation of $M$. }
\end{figure*}

 In this paper, we assume $\phi $ is known for the manifold. The Laplace-Beltrami operator in \eqref{eqn:heat} is the infinitesimal generator of BM on manifolds \cite{hsu1988brownian}. Simulating sample paths of BM on $M$ with starting point $s_0$ is equivalent to simulating a stochastic process in $R^d$ with starting point $x_0$. An example of BM on torus is given in Fig.\ref{fig:bm}. The right panel illustrates the plotting of three paths of a stochastic process. These paths are the BM paths on a torus, obtained by mapping the stochastic process to the torus using $\phi$. The BM on a Riemannian manifold in a local coordinate system is given as a system of stochastic differential equations (SDE) in the Ito form \cite{hsu1988brownian,hsu2008brief}:
\begin{equation}
\label{eqn:bmsde}
d x_{i}(t)=\frac{1}{2} G^{-1 / 2} \sum_{j=1}^{d} \frac{\partial}{\partial x_{j}}\left(g^{i j} G^{1 / 2}\right) d t+\left(g^{-1 / 2} d B(t)\right)_{i},
\end{equation}
where $g$ is the metric tensor of $M$, $B(t)$ represents a BM in the Euclidean space, $G$ is the determinant of $g$.  If $M$ is $R^d$, $g$ becomes an identity matrix and $x_{i}(t)$ is the standard BM in $R^{d}$.  We provide an example of how to derive the metric tensor and the SDE of BM on Bitten Torus in Appendix A. 

\section{In-BO on manifolds}

In-BO is a search strategy for finding the global optimal on complex constrained domains. It extends the Tra-BO algorithm to manifold-valued data. There are two phases for In-BO. In the initial phase, we simulate BM sample paths on the manifold and estimate the heat kernel by evaluating the transition density from the BM paths. In the iterative phase, the Sparse In-GP  surrogate model is constructed using the heat kernel on manifolds. Its predictive mean and variance are used in the acquisition function to sample the next `best point' $s_i$ from the grid points.
In every iteration, the next `best point' found from the acquisition function in the last iteration will be added into the training set 
\begin{align*}
\label{eqn:trainset}
\mathcal{D}_{1:i}= \{ \mathcal{D}_{1:i-1}, (\mathbf{s}_{i}, \mathbf{y}_i)\} 
\end{align*}
where $\mathbf{y}_i$ is the observations of the objective function at $\mathbf{s}_i$. The augmented training set will be used to update the posterior of SIn-GP. 

\subsection{The surrogate model: Sparse In-GP on manifolds }

It is not straightforward to define a valid covariance kernel for GP when the predictors are on the manifold. Although theoretically a heat kernel is a natural choice, it is analytically intractable to directly evaluate. We will utilize the relationship between heat kernels and the transition density of BM on manifolds for constructing covariance kernels. The numerical approximation of the heat kernel $K_{\text {heat }}^{t}$ can be derived as below. We define $\mathrm{S}(\mathrm{t})$ as a BM on $M$ starting from $s_{0}$ at time $t = 0$. The probability of $\mathrm{S}(\mathrm{t}) \in \mathrm{A} \subset \mathrm{M}$, for any Borel set $A$, is given by
\begin{equation}
\mathbb{P}\left[S(t) \in A \mid S(0)=s_{0}\right]=\int_{A} K_{\text {heat }}^{t}\left(s_{0}, s\right) d s,
\end{equation}
where the integral is defined as the volume form of $M$. The transition probability is approximated as
\begin{equation}
\mathbb{P}\left[S(t) \in A \mid S(0)=s_{0}\right] \approx \frac{k}{N},
\end{equation}
where $N$ is the total number of simulated BM sample paths and $k$ is the number of BM paths which reach $A$ at time $t$. Figure 2 shows three BM sample paths starting from same initial point on the surface of a Bitten Torus.  Only the green sample path reaches $A$ at time $t$, so the estimate of the transition probability $p\left(S(t) \in A \mid S(0)=s_{0}\right)$ is 1/3.

The transition density of $S(t)$ at $s$ is approximated as:
\begin{equation}
\label{eq:bmden}
K_{h e a t}^{t}\left(s_{0}, s\right) \approx \hat{K}^{t}=\mathbb{P}\left[S(t) \in A \mid S(0)=s_{0}\right] \approx \frac{1}{V(A)} \cdot \frac{k}{N},
\end{equation}
where $V(A)$ is the Riemannian volume of $A$, and $\widehat{K}^{t}$ is the estimated transition density, as well as the estimated heat kernel. The Neumann boundary condition corresponds to BM reflecting at the boundary. This can be approximated by pausing time and resampling the next step until it falls into the interior of $M$. This estimator is asymptotically unbiased and consistent \cite{niu2019intrinsic}. Note that $t$ is the BM diffusion time. If $t$ is large, the BM paths have a higher probability to reach $A$.

Let $\Sigma$ be the covariance matrix for all grid points on $M$. Given two grid points $s_i$ and $s_j$ on $M$, $\Sigma_{i,j}$ can be constructed by simulating $N$ BM paths starting at $s_i$ and numerically evaluating the transition density of the BM at $s_j$ using \eqref{eq:bmden}. We have



\begin{equation}
\label{eqn:kernelM}
\Sigma_{i j}=\sigma_{h}^{2} K_{h e a t}^{t}\left(s_{i}, s_{j}\right).
\end{equation}

We introduce the rescaling hyperparameter $\sigma_h^2$ to add extra flexibility to control the magnitude of the kernel. Once the $N$ BM paths are simulated, the heat kernel between $s_i$ and other grid points can be evaluated from the existing simulated paths. 

Under the GP prior for the unknown objective function $f: M \rightarrow R$, we have: 
\begin{equation}
\label{eq:prior}
p(\mathbf{f} | s \in S ) = \mathcal{N}( 0 , \Sigma),
\end{equation}
where $\mathbf{f}$ is the vector of the realisation of $f$ at the grid points.
Let $\mathbf{f}_\mathcal{D}$ be the observed data point in the BO training set $\mathcal{D}$ and $\mathbf{f}_r$ be the vector of $f(.)$ at the grid points which is not in $\mathcal{D}$. The joint distribution of $\mathbf{f}_\mathcal{D}$ and $\mathbf{f}_r$ is :

\begin{equation}
p(\mathbf{f}_\mathcal{D}, \mathbf{f}_{r})=\mathcal{N}\left(0,\left[\begin{array}{ll}
\Sigma_{\mathrm{\mathcal{D}\mathcal{D}}} & \Sigma_{\mathrm{\mathcal{D}r}} \\
\Sigma_{\mathrm{r\mathcal{D}}} & \Sigma_{\mathrm{rr} }
\end{array}\right]\right)\,
\end{equation}
where $\Sigma_{\mathcal{D}\mathcal{D}}$ is the covariance matrix for all data points in $\mathcal{D}$ and $\Sigma_{rr}$ is the covariance matrix for all grid points not in $\mathcal{D}$.
To construct the matrix $\Sigma$ for all grid points, we need to simulate BM paths for every grid point. Although the BM simulations are trivially parallelizable, the computational cost can be high when the number of grid points is large. To solve this problem, we propose the sparse In-GP on manifolds by introducing inducing points. The GP prior can be augmented with $m$ inducing points on $M$ denoted as $\bm{z}=[z_{1},z_{2},...,z_{m}], z_{i}\in M$. In our work, the inducing points is a subset of the grid points. The realisation of the objective function at the inducing points can be represented as the vector $\bm{u}=\left[f\left(z_{1}\right), \ldots, f\left(z_{m}\right)\right]$. The prior distribution can be written in terms of $p(\bm{u})$ and the conditional distribution $p(\mathbf{f} | \bm{u} )$.
\begin{equation}
p(\mathbf{f}) = \int p(\mathbf{f} | \bm{u}) p(\bm{u}) d\bm{u}, \ \ \ p(\bm{u}) = \mathcal{N}( 0 , \Sigma_{zz} ),
\end{equation}
where $\Sigma_{zz}$ is the covariance matrix for all inducing points. By using the deterministic inducing conditional approximation in \citep{quinonero2007approximation}, $\mathbf{f}_\mathcal{D}$ and $\mathbf{f}_r$ are assumed to be conditionally independent given $\mathbf{u}$. The GP prior can be approximated by the SIn-GP prior as

\begin{equation}
\begin{aligned}
p(\mathbf{f}_\mathcal{D}, \mathbf{f}_r ) &\approx q\left(\mathbf{f}_\mathcal{D}, \mathbf{f}_{r}\right) \\
q\left(\mathbf{f}_\mathcal{D}, \mathbf{f}_{r}\right)&=\mathcal{N}\left(0,\left[\begin{array}{ll}
Q_{\mathrm{\mathcal{D}\mathcal{D}}} & Q_{\mathrm{\mathcal{D}r} .} \\
Q_{\mathrm{r\mathcal{D}}} & Q_{\mathrm{rr} }
\end{array}\right]\right)\\
&=\mathcal{N}\left(0,\left[\begin{array}{cc}
\Sigma_{\mathrm{\mathcal{D}z}} \Sigma_{\mathrm{zz}}^{-1} \Sigma_{\mathrm{z\mathcal{D}}} & \Sigma_{\mathrm{\mathcal{D}z}} \Sigma_{\mathrm{zz}}^{-1} \Sigma_{\mathrm{zr} }\\
\Sigma_{\mathrm{rz}} \Sigma_{\mathrm{zz}^{-1} \Sigma_{\mathrm{z\mathcal{D}}}} & \Sigma_{\mathrm{rz}} \Sigma^{-1}_{ \mathrm{zz}}
\Sigma_{\mathrm{zr}}
\end{array}\right]\right),
\end{aligned}
\end{equation}
where $\Sigma_{\mathrm{zz}}$, $\Sigma_{\mathrm{\mathcal{D}z}}$, and $\Sigma_{\mathrm{zr}}$ can all be obtained by evaluating the transition densities from the BM simulations with inducing points as the starting points. The number of inducing points is much smaller than the number of data points. BM paths only need to be simulated starting at the inducing points instead of every data point. This approximation summarizes the training data into a small set of inducing points, so that inference could be done more efficiently \cite{quinonero2007approximation}. 

Let $\mathbf{y}$ be the observation of the objective function in the training set $\mathcal{D}$. With this approximation the approximated marginal likelihood can be written as:
\begin{equation}
p(\mathbf{y}) \approx q(\mathbf{y} ) = \mathcal{N} (0 , \Sigma_{\mathrm{\mathcal{D}z}} \Sigma_{\mathrm{zz}}^{-1} \Sigma_{\mathrm{z\mathcal{D}}} + \sigma_{noise}^2 \mathbf{I} ).
\end{equation}

The diffusion time $t$ and the magnitude parameter $\sigma^2_h$ can be optimised as the hyperparameter of the kernel by maximising this approximated likelihood.
The predictive distribution is shown below:
\begin{equation}
\label{eqn:predict}
\begin{aligned}
q\left(\mathbf{f}_{r} \mid \mathbf{y}\right)=\mathbf{N}\left(Q_{\mathrm{r\mathcal{D} }}\left(Q_{\mathrm{\mathcal{D}\mathcal{D}}}+\sigma_{noise}^{2} \mathbf{I}\right)^{-1} \mathbf{y}\right.,\\
\left.Q_{\mathrm{rr}}-Q_{\mathrm{r\mathcal{D} }}\left(Q_{\mathrm{\mathcal{D}\mathcal{D}}}+\sigma_{noise}^{2} \mathbf{I}\right)^{-1} Q_{\mathrm{\mathcal{D}r } }\right).
\end{aligned}
\end{equation}




The BM simulations only need to be run once for the inducing points before the subsequent optimisation steps.
When new observations are added in the BO training set $\mathcal{D}$, the new $\Sigma_\mathrm{z\mathcal{D}}$ can be constructed based on the existing BM simulations. 


\subsection{The Acquisition Function-Probability Improvement}

The acquisition function is used to determine the next optimal location from the grid points to evaluate the objecting function during the optimization process. 
It is constructed based on the posterior distribution obtained from the observed data set $\mathcal{D}_{i}$, and defined as $\alpha\left(\boldsymbol{\cdot } | \mathcal{D}_{i}\right)$.
The next evaluation point $s_{i+1}$ is selected from the grid points by maximizing:
\begin{equation}
\boldsymbol{s}_{i+1}=\operatorname{argmax} _{\boldsymbol{s} \in S} \alpha\left(s  | \mathcal{D}_{i}\right).
\end{equation}
The acquisition function takes into account both the predictive mean and  variance of the probabilistic surrogate model to balance exploration and exploitation in the search for the optimal solution. There are many different types of acquisition functions. Improvement-based acquisition functions are designed to identify points in the search space where the objective function is likely to improve upon the current best known solution. In this research, we select PI as the acquisition function which can be expressed as:
\begin{equation}
\label{PI}
s_{i+1}=\operatorname{argmax}_{s} \Phi\left(\frac{\mu_{i}(s)-f\left(s^{+}\right)-\epsilon}{\sigma_{i}(s)}\right),
\end{equation}
where $s^{+}=\operatorname{argmax}_{s_{i} \in s_{1 : i}} f\left(s_{i}\right)$, $s^{+}$ represents the position where the objective function f is maximized after obtaining $i^{th}$ sample points; $s_{i}$ is the location to be found in step i; $\Phi$ represents the cumulative distribution function of Gaussian distribution; $\mu _{i}(s)$ is the predictive mean of SIn-GP while $\sigma_{i}(s)$ is the predictive standard deviation of SIn-GP shown in \eqref{eqn:predict}. $\epsilon$ controls the degree of exploration in PI, balancing the relationship between local and global search. The most `potential optimal' evaluation point is selected by maximising \eqref{PI}.

We summarize In-BO in Algorithm 1. The initial phase begin with selecting inducing points on the manifold. In this work, the inducing points are selected from the grid points which are equally spaced on the manifold. The BM paths are simulated using \eqref{eqn:bmsde} starting from the inducing point. The heat kernel is estimated as the BM transition density using \eqref{eq:bmden}. The covariance matrices of the inducing points and all grid points can be constructed using the heat kernel estimates at different diffusion time.  This concludes the initial phase. In the iterative phase, the training set $\mathcal{D}$ is initialised by randomly choosing initial locations from the grid points. The PI acquisition function uses the predictive mean and variance from SIn-GP to search for the next `best location', $s_{i+1}$. In every iteration, $s_{i+1}$ is added into training set $\mathcal{D}$ to update the posterior distribution of SIn-GP. The updating process is repeated until the maximum number of iterations is reached. Throughout the process, the BM paths only need to be simulated once in the initial phase.


\begin{algorithm}[tb]
   \caption{ Intrinsic Bayesian Optimisation on manifolds}
   \label{alg:bm}
\begin{algorithmic}
   \STATE {\bf Initial Phase}
   \item 1.1 Select $n$ inducing points from the grid points on $M$;
   \item 1.2 Simulate BM paths starting with inducing points:
   \FOR{$i = 1,\ldots,m$   \COMMENT{ \footnotesize $m$ is the size of inducing points } }
   \FOR{$j = 1,\ldots,N$   \COMMENT{\footnotesize $N$ is No. of paths }  }
   \STATE Simulate BM sample paths starting at $ith$ inducing point using \eqref{eqn:bmsde};
     \ENDFOR
    \ENDFOR
    \item 1.3. Estimate the transition density of BM on $M$ using \eqref{eq:bmden};
   \item 1.4 Construct $\Sigma_{zz}, \Sigma_{zr}$ using the heat kernel as in \eqref{eqn:kernelM};
\STATE
\STATE {\bf Iterative Phase} 
\STATE 2.1 Initialise the In-BO by selecting the initial locations from the grid points on $M$. $\mathcal{D}_0$ is initialised as $\mathcal{D}_0 = \{ \mathbf{s}_0 , \mathbf{y}_0 \} $
\FOR{$i = 1,\ldots,I$  \COMMENT{ \footnotesize $I$ is the number of iterations } }
\item 1. Calculate predict mean and predict variance on grid points from SIn-GP, using \eqref{eqn:predict}; $\Sigma_{z\mathcal{D}}$ can be constructed from $\Sigma_{zr}$ by selecting the corresponding rows and columns;
\item 2. Find $s_{i+1}$ by optimizing PI function as in \eqref{PI};
\item 3. The training set is augmented $\mathcal{D}_i = \{ \mathcal{D}_{i-1}, (  \mathbf{s}_i, \mathbf{y}_i ) \} $;
   \ENDFOR
\end{algorithmic}
\end{algorithm}

\section{Application of In-BO on manifolds}
The applications of In-BO are illustrated in the simulation studies on the U-shape domain, the Bittern Torus and the real-world dataset of the Aral sea.


\subsection{U-shape Application}
We first apply the In-BO to the U-shape domain, where the value of the function varies smoothly from the lower right corner to the upper right corner of the domain ranging from -6.19 to 6.19, shown in Figure 3(a). In our simulation study, we have 285 grid data points equally spaced within the boundary. The orange star marks the location of the maximum function value in grid points. Since the U-shape domain is a subset of $R^{2}$, we only need to run standard BM starting from the inducing points in the two dimensional Euclidean space restricted within the U-shape boundary. The BM paths are simulated for 20 inducing points which are equally spaced within the boundary.

With PI as the acquisition function, 3 points are randomly picked from the grid points as the initial locations of the optimisation. we show optimisation results for the In-BO and Tra-BO separately in Figure 3(b-e). The results of the Tra-BO is very sensitive to the initial location from where the optimization starts. If the initial points contain one or more points from the upper side, Tra-BO can leverage this information to guide its search towards the global optimum. As shown in Figure 3(b), Tra-BO successfully finds the optimal point, marked by a blue star. When the initial points are all located in the lower side of the U-shape and no information about the upper side is provided, Tra-BO is unable to find the optimal point. It will be influenced by the information obtained from initial locations and stay in the middle part or to the left, as shown in Figure 3(c).
In-BO can find the the global maximum in the U-shape domain in both cases, as shown in Figures 3(d), 3(e). Tra-BO is sensitive to the initial location, while In-BO is much more robust. 
\begin{figure*}[ht]
	\centering
 \subfigure[True function]{
		\begin{minipage}[0.3]{0.27\textwidth}
			\includegraphics[width=1\textwidth]{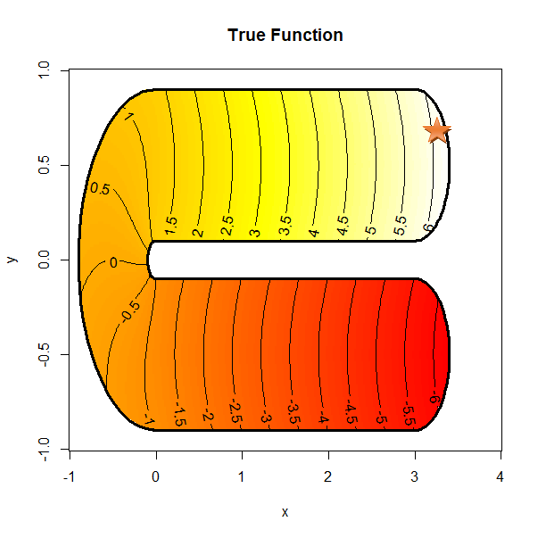} 
		\end{minipage}
		\label{fig:grid_4figs_1cap_4subcap_1}
	}
	\subfigure[Tra-BO initialised from upper side]{
		\begin{minipage}[0.3]{0.27\textwidth}
			\includegraphics[width=1\textwidth]{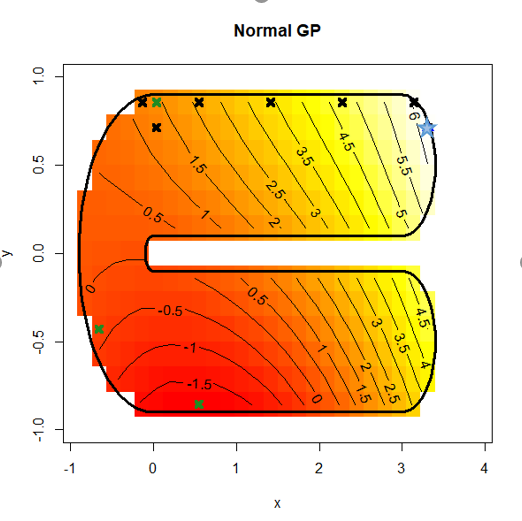} 
		\end{minipage}
		\label{fig:grid_4figs_1cap_4subcap_1}
	}
    	\subfigure[Tra-BO initialised from lower side]{
    		\begin{minipage}[0.3]{0.27\textwidth}
   		 	\includegraphics[width=1\textwidth]{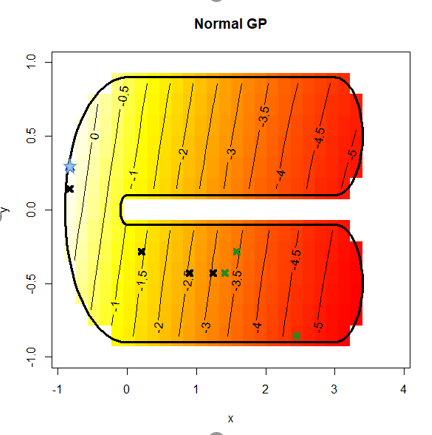}
    		\end{minipage}
		\label{fig:grid_4figs_1cap_4subcap_2}
    	}
	\subfigure[In-BO initialised from upper side]{
		\begin{minipage}[0.3]{0.27\textwidth}
			\includegraphics[width=1\textwidth]{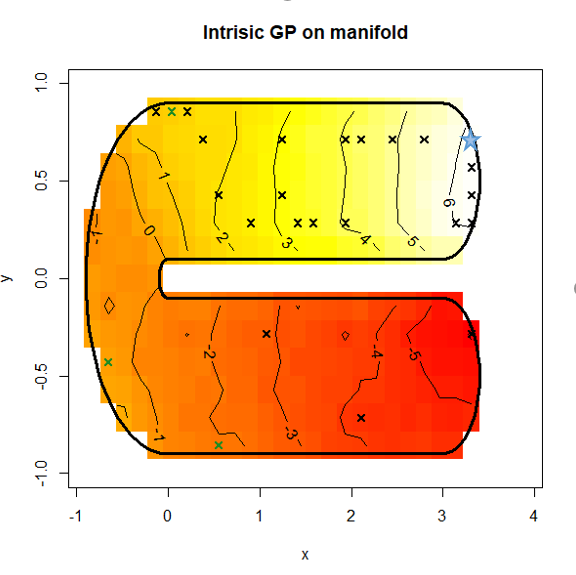} 
		\end{minipage}
		\label{fig:grid_4figs_1cap_4subcap_3}
	}
    	\subfigure[In-BO initialised from lower side ]{
    		\begin{minipage}[0.3]{0.27\textwidth}
		 	\includegraphics[width=1\textwidth]{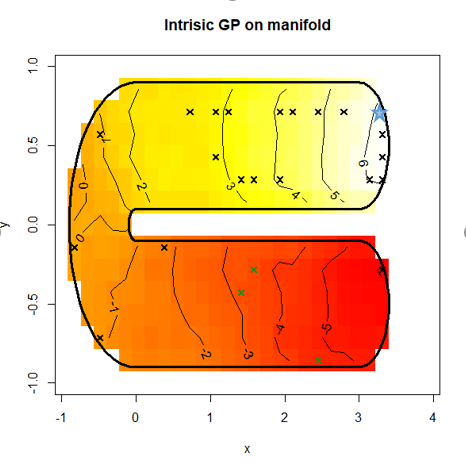}
    		\end{minipage}
		\label{fig:grid_4figs_1cap_4subcap_4}
    	}
     \subfigure[Violin plot]{
		\begin{minipage}[0.3]{0.27\textwidth}
			\includegraphics[width=1\textwidth]{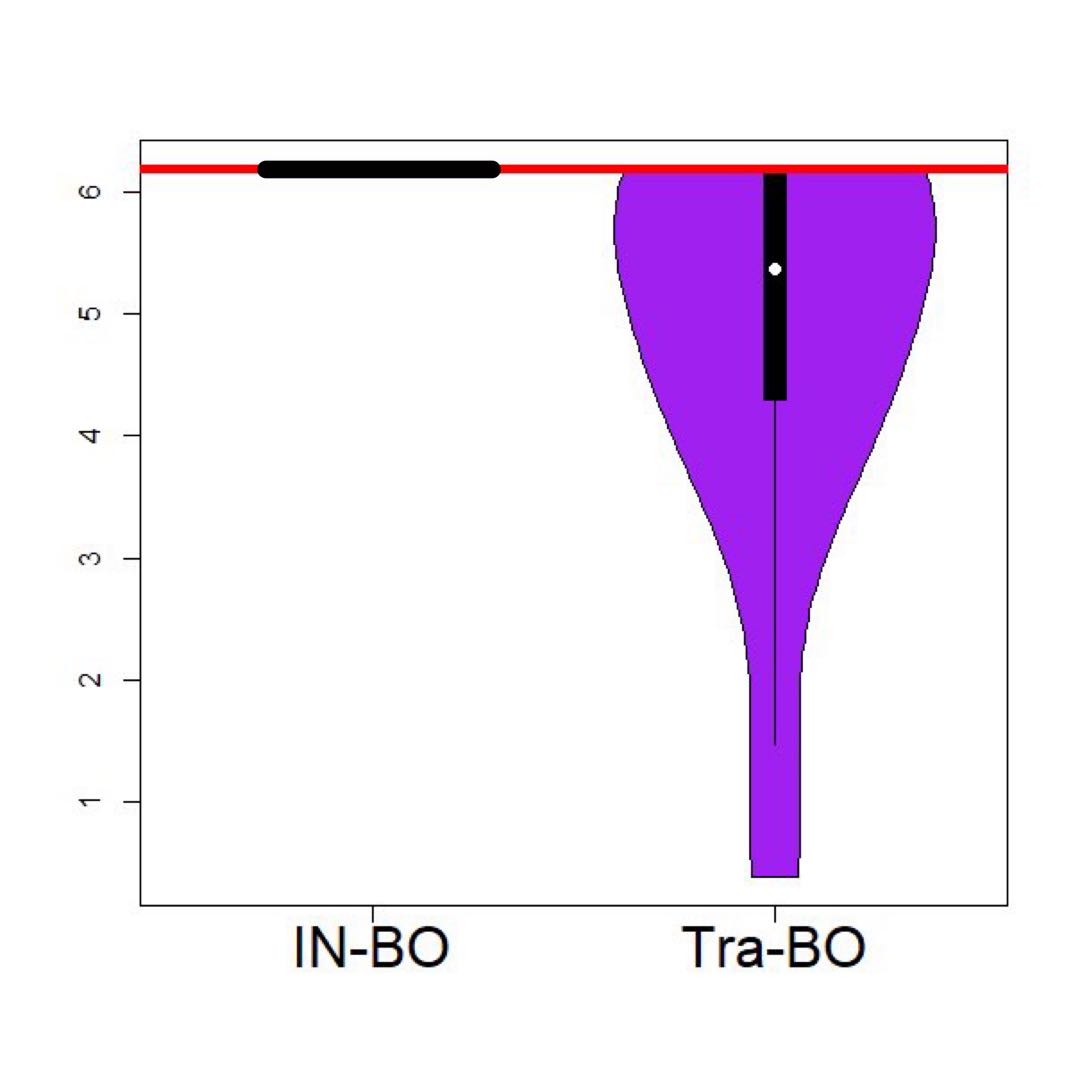} 
		\end{minipage}
		\label{fig:grid_4figs_1cap_4subcap_1}
	}
	\caption{(a): The true function is plotted in the U-shape domain; the orange star shows the location with maximum value. (b): The Tra-BO optimisation steps when the initial locations are in the upper side. The green crosses are the initial points, and the black crosses are the optimal location in each iterations, and the blue star is the global optimal found by Tra-BO. (c): Tra-BO optimisation steps when the initial locations are in the lower side. (d): In-BO optimisation steps using the same initial locations as in (b). (e): In-BO optimisation steps using the same location as (c). (f): The violin plot of the In-BO and Tra-BO, generated from 20 sets of random initialisation. The red horizontal line indicates the true global maximum.}
	\label{fig:grid_4figs_1cap_4subcap}
\end{figure*}

We randomly select 20 sets of initial locations from grid points. Each set contains 3 initial locations. For each set of the initialisation, the In-BO and Tra-BO are applied to search for the maximum value of the objective function in the U-shape domain. The results are plotted in the violin plots in Figure 3(f). The red horizontal line indicate the true global optimal value. The black line indicate the maximum value found by In-BO for each initialisation. It is clear that the In-BO can find the true global maximum regardless of the initialisation. The distribution of Tra-BO results is plotted in purple. The performance of Tra-Bo is very sensitive to the choice of initialisation. It can only find the true global maximum when the initial locations appear in the upper side of the domain.



\subsection{Bitten Torus Application}
Bitten torus is a two-dimensional manifold embedded in $\mathbb{R}^{3}$. The construction of Bitten Torus is given in Appendix A.1. 600 grid data points are equally spaced on the surface of the Bitten torus. 
The value of the objective function increases smoothly from 0.57 to 5.50 on the surface of the Bitten Torus, which can be seen from the colour plot in Figure 6(a), where dark blue represents low values and bright red represents high values. The two ends of the bitten torus (dark blue and bright red) have a small Euclidean  distance. 

The BM paths need to be simulated to estimate the heat kernel on Bitter Torus before the optimization process start. We set 19 inducing points which are equally distributed on the surface of the Bitten Torus. The BM sample paths on the manifold can be simulated using \eqref{eqn:bmsde} starting from the inducing points. The metric tensor of the Bitten Torus is derived in Appendix A.1. The BM on Bitten Torus can be constructed via the stochastic differential equations:
\begin{equation}
\begin{aligned}
d \theta(t) & =-\frac{1}{2} r^{-1} \sin \theta(R+r \cos \theta)^{-1} d t+r^{-1} d B_1(t) \\
d \phi(t) & =\left|(R+r \cos \theta)^{-1}\right| d B_2(t)
\end{aligned}.
\end{equation}
where $R$ and $r$ are fixed, $\theta$ represents the angle of torus and $\phi$ represents angle of tube. The detailed derivation process can be viewed in the Appendix A.
An example of the optimisation steps in Tra-BO is shown in Figure 4(b). When initial points are located in the dark blue area, the Tra-BO is misguided by the Euclidean distance based surrogate model. The function values at the two ends of the Bitten Torus are predicted to be similar subject to the Euclidean distance. For the same initialisation, the optimisation steps of the In-BO is shown in Figure 4(c). With the help of the SIn-GP on manifolds, the In-BO can take into account the geometry of the Bittern torus and find the global maximum. 



We randomly select 20 sets of initial locations from grid points on the Bittern Torus. Each set contains 4 initial locations. Under each set of initialisation, In-BO and Tra-BO are applied to search for the maximum value of the objective function. Results are shown in the violin plots in Figure 4(d). The red horizontal line indicate the true global optimal value. The black line indicate the maximum value found by In-BO for each initialisation. The distribution of Tra-BO results is plotted in purple. It is clear that the In-BO can find the true global maximum for all different initialisation set, while the Tra-BO is very sensitive to the choice of the initial location.



\begin{figure*}[!t]
	\centering
 \subfigure[True function]{
		\begin{minipage}[b]{0.215\textwidth}
			\includegraphics[width=1\textwidth]{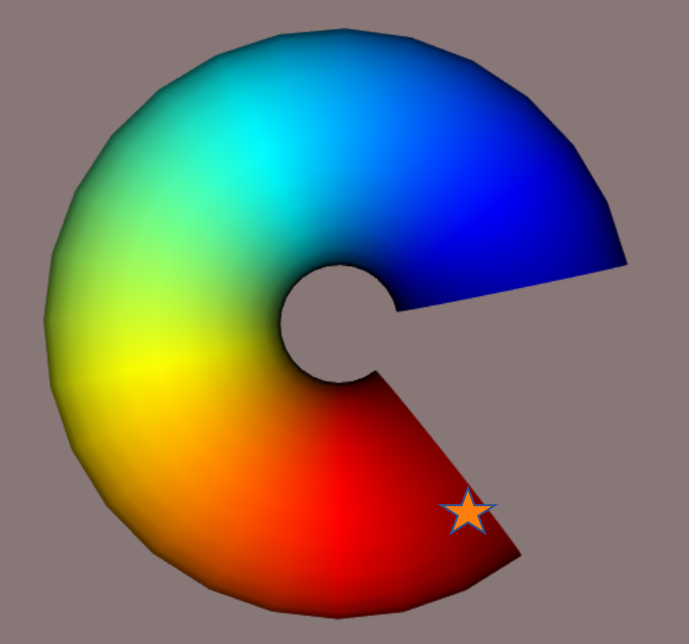} 
		\end{minipage}
		\label{fig:grid_4figs_1cap_4subcap_1}
	}
	\subfigure[Tra-BO initialised from dark blue part]{
		\begin{minipage}[b]{0.24\textwidth}
			\includegraphics[width=1\textwidth]{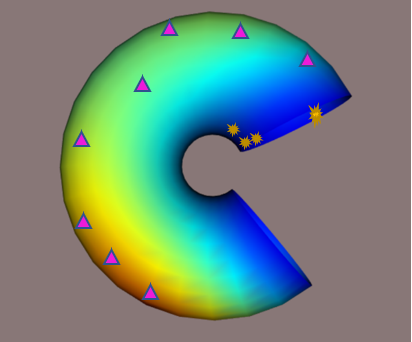} 
		\end{minipage}
		\label{fig:grid_4figs_1cap_4subcap_1}
	}
    	\subfigure[In-BO initialised from dark blue part]{
    		\begin{minipage}[b]{0.215\textwidth}
   		 	\includegraphics[width=1\textwidth]{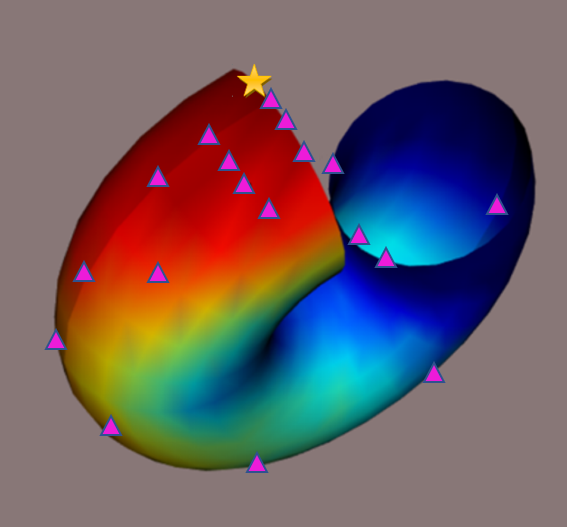}
    		\end{minipage}
		\label{fig:grid_4figs_1cap_4subcap_2}
    	}
     \subfigure[Violin plot]{
    		\begin{minipage}[b]{0.245\textwidth}
   		 	\includegraphics[width=1\textwidth]{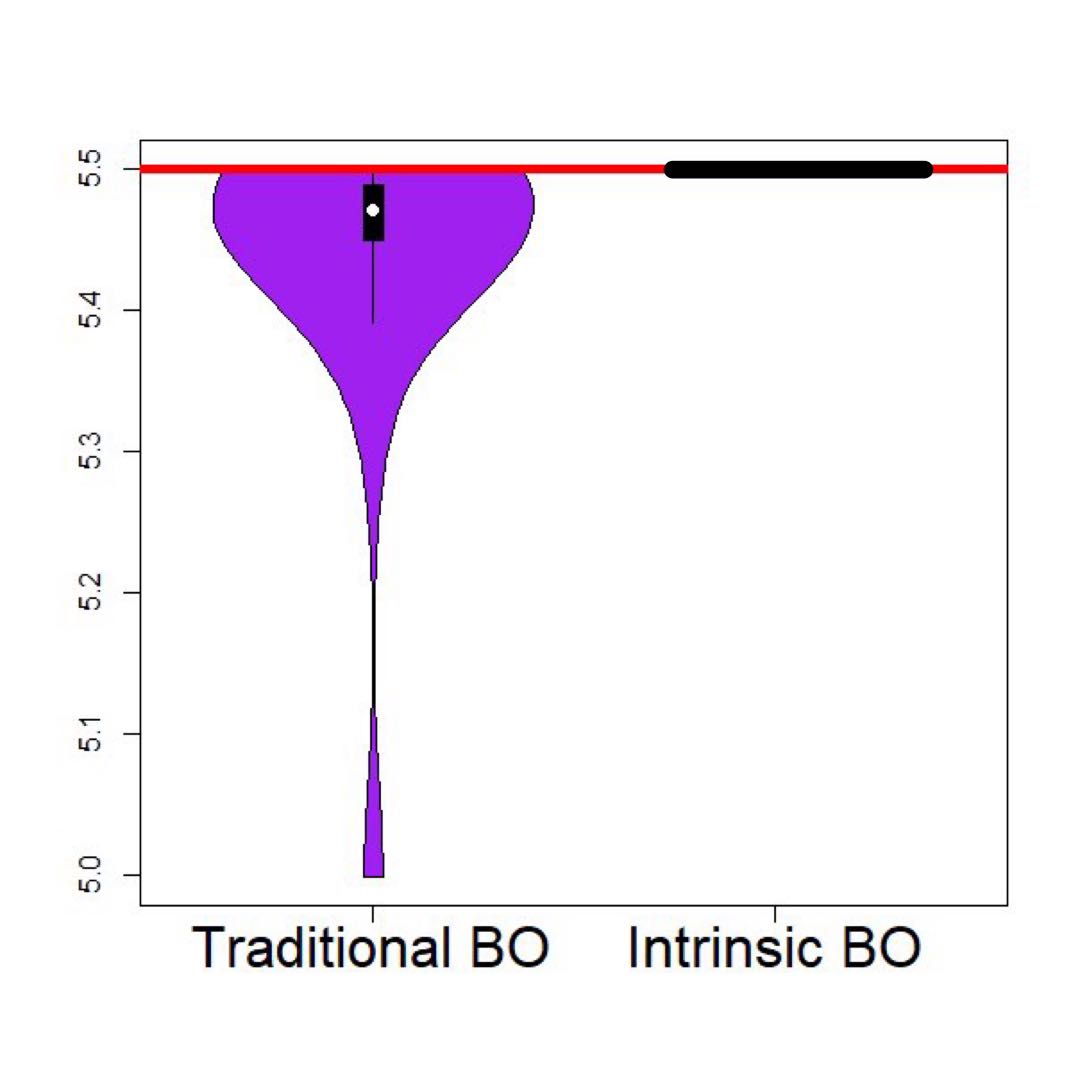}
    		\end{minipage}
		\label{fig:grid_4figs_1cap_4subcap_2}
    	}
	\caption{(a): The plot of true function on Bitten Torus; the orange star represents the location with the largest value. (b): Tra-BO optimisation steps when the initial locations are in the blue area of the Bitten Torus. (c): In-BO optimisation steps  using the same initial locations as (b). (d): The violin plot of the In-BO and Tra-BO, generated from 20 sets of random initialisation.
  The red horizontal line indicates the true global maximum.}
	\label{fig:grid_4figs_1cap_4subcap}
\end{figure*}

\subsection{Aral sea Application}
We consider remotely sensed chlorophyll data in the Aral Sea to investigate sites with the highest chlorophyll concentration. Figure 5(a) plots measurements of chlorophyll levels in the Aral sea \cite{wood2008soap}. The level of chlorophyll concentration is represented by the intensity of the colour. In the Aral Sea, there are a total of 485 grid data points scattered within the boundary. The location with highest level of chlorophyll concentration is shown as blue star in Figure 5(a).
BM paths are simulated to estimate the heat kernel of the domain. We select 42 inducing points from the grid points which are equidistantly distributed within the boundaries of the Aral Sea. 

In conducting corresponding inferences and prediction tasks, it is important to take into account the intrinsic geometry of the sea and its complex boundary. The Euclidean distance between two points is the `straight-line' distance between them, which is not always an accurate measure of the actual distance between them when taking into account factors such as a land barrier. These locations have quite different chlorophyll levels. The chlorophyll data are noisy, varying within the boundary but not across the gap. Tra-BO do not consider the boundary and would naturally provide close estimates of the chlorophyll level given their close spatial vicinity, which can be seen from Figure 5(b). the surrogate model of the Tra-BO smooths across the isthmus of the central peninsula. However, with the same initial locations as in Figure 5(b), the In-BO performs better than Tra-BO, which can be seen from Figure 5(c).

We randomly select 20 sets of initial locations from 485 grid points to implement Tra-BO and In-BO. Each set contains 4 initial points. Results are shown by the violin plots in Figure 5(d). Due to the noisiness and the nonstationary characteristics of the  chlorophyll data, the optimal point cannot be find for every set of initialisation. However, it can still be seen from the violin plot that In-BO outperforms Tra-BO. More results of this application can be obtained from Appendix A.2.
\begin{figure*}[!htb]
	\centering
 \subfigure[True function]{
		\begin{minipage}[b]{0.22\textwidth}
			\includegraphics[width=1\textwidth]{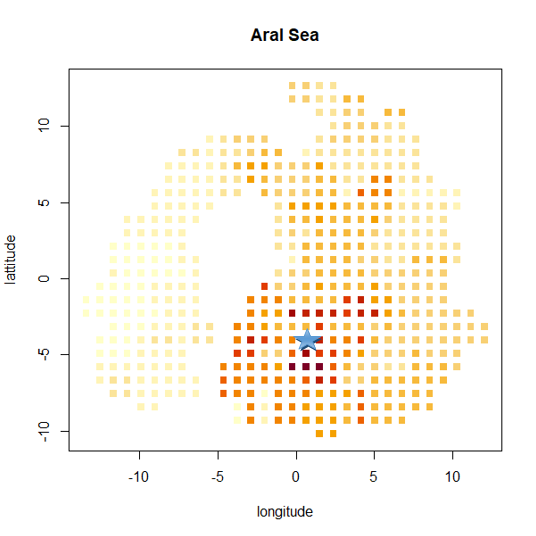} 
		\end{minipage}
		\label{fig:grid_4figs_1cap_4subcap_1}
	}
  \subfigure[Tra-BO optimisation steps ]{
		\begin{minipage}[b]{0.22\textwidth}
			\includegraphics[width=1\textwidth]{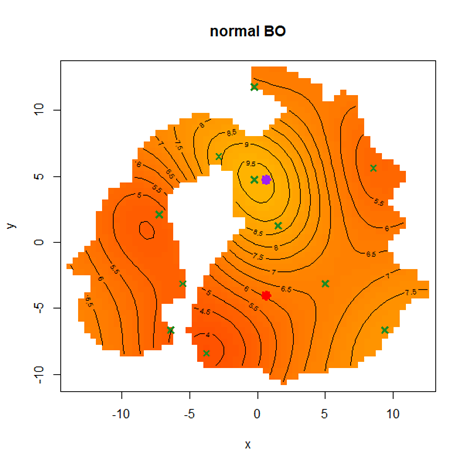} 
		\end{minipage}
		\label{fig:grid_4figs_1cap_4subcap_1}
	}
	\subfigure[In-BO optimisation steps using same initial locations as (b)]{
		\begin{minipage}[b]{0.22\textwidth}
			\includegraphics[width=1\textwidth]{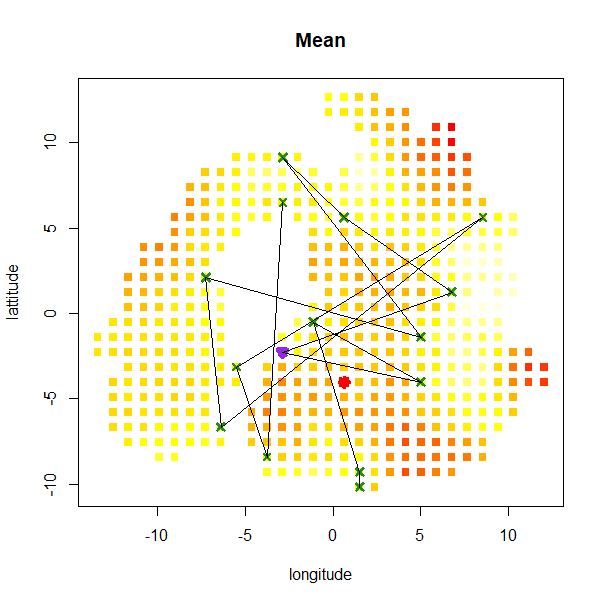} 
		\end{minipage}
		\label{fig:grid_4figs_1cap_4subcap_1}
	}
    	\subfigure[Violin plot]{
    		\begin{minipage}[b]{0.22\textwidth}
   		 	\includegraphics[width=1\textwidth]{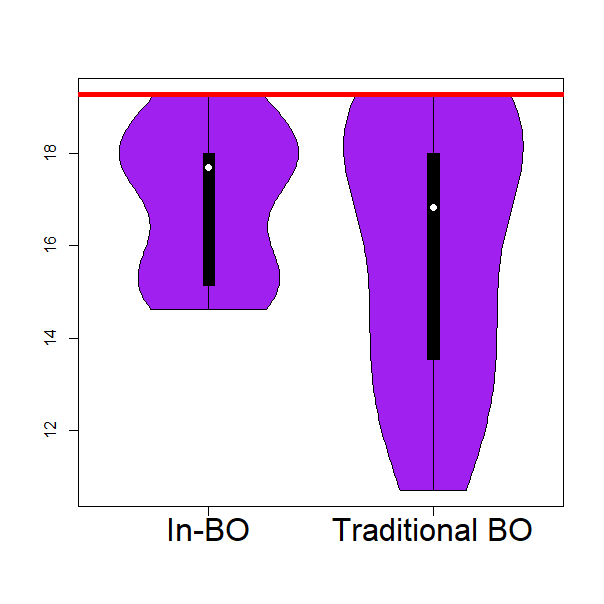}
    		\end{minipage}
		\label{fig:grid_4figs_1cap_4subcap_2}
    	}
	\caption{(a): The chlorophyll levels in the Aral sea; the distribution of the chlorophyll levels in the west and east side of the sea are quite different due to the land barrier; The blue star represents the location with the highest chlorophyll concentration. (b): Tra-BO optimisation steps when the initial locations are close to the boundary. The contour represent the predictive mean of the Euclidean distance based surrogate model. It smooths across the boundary. The green crosses are the training dataset. The red dot represents the location with the highest chlorophyll level and the purple dot represents the optimal point found by Tra-BO. (c): In-BO optimisation steps with the same initial locations as in (b). (d): The violin plot of the In-BO and Tra-BO: with 20 sets of initial points randomly selecting from grid points. The red horizontal line indicates the true global maximum.}
	\label{fig:grid_4figs_1cap_4subcap}
\end{figure*}

\section{Discussion and conclusion}
Our work proposes a general methodology of In-BO on manifolds and complex constrained domains based on the SIn-GP surrogate model using the heat kernel on manifolds. Obtaining knowledge of the geometry and intricate boundaries of the manifold through the heat kernel allows BO to more effectively navigate the manifold and enhance the precision of resolving the optimization problem. In-BO is applied on different examples, including U-shape domain, Bitten Torus and Aral sea. When the implicit optimization objective function is smooth such as the objective function in U-shape domain and Bitten Torus, In-BO consistently discovers the global optimum regardless of the initialization. In contrast, the Tra-BO is highly sensitive to the initialization. It can only locate the global optimum when the initial position is very close to it, otherwise it will get stuck in a local optimum. When the implicit objective function is non-differentiable, multi-peaked and non-convex, In-BO still presents outstanding advantages compared to Tra-BO in the Aral sea example. In-BO is much more robust while Tra-BO is sensitive to the initial locations. There has been abundant interest in learning of In-BO on unknown manifolds, which means no intrinsic geometry available in advance. In future work, we will explore the application of In-BO in unknown manifolds.

\bibliography{example_paper}
\bibliographystyle{icml2022}
\newpage
\appendix
\section{Appendix}
\subsection{Brownian motion on the Bitten Torus}
The Geometric Brownian Motion on a Riemannian manifold is given as a system of stochastic
differential equations, refer to \cite{hsu2008brief}\cite{hsu1988brownian}:
\begin{equation}
d x_i(t)=\frac{1}{2} G^{-1 / 2} \sum_{j=1}^d \frac{\partial}{\partial x_j}\left(g^{i j} G^{1 / 2}\right) d t+\left(g^{-1 / 2} d B(t)\right)_i,
\end{equation}
where $g$ is the metric tensor of Riemannian manifold $M$, $g^{i j}$ is the $(i, j)$ element of its inverse, $G$ is the determinant of the matrix $g$ and $B(t)$ represents an independent Brownian Motion in the Euclidean space.
Using the Euler-Maruyama method\cite{kloeden1992higher}\cite{lamberton2011introduction}, the equation can be derived in:
\begin{equation}
\begin{aligned}
d x_i(t)=\frac{1}{2} \sum_{j=1}^2\left(-g^{-1} \frac{\partial g}{\partial x_j(t)} g^{-1}\right)_{i j} d t
+\frac{1}{4} \sum_{j=1}^2\left(g^{-1}\right)_{i j}\\
\operatorname{tr}\left(g^{-1} \frac{\partial g}{\partial x_j(t)}\right) d t+\left(g^{-1 / 2} d B(t)\right)_i
\end{aligned}.
\end{equation}
For the Bitten Torus, we parametrise the three-dimensional coordinates of it by 4 variables:
$r$ radius of tube, $R$ distance from centre of the tube to the centre of the torus, $\theta $ and $\phi $ are angles to make full circles while $\theta $ for angle of torus and $\phi $ for angle of tube. The Torus can be expressed as: 
\begin{equation}
\mathbf{X}(\theta, \phi)=((R+r \cos \theta) \cos \phi,(R+r \cos \theta) \sin \phi, r \sin \theta).
\end{equation}
We first compute the partial derivatives for $\theta $ and $\phi $:
\begin{equation}
\begin{aligned}
\mathbf{X}_\phi & =((R+r \cos \theta)(-\sin \phi),(R+r \cos \theta) \cos \phi, 0),\\
\mathbf{X}_\theta & =(r \cos \phi(-\sin \theta), r \sin \phi(-\sin \theta), r \cos \theta)
\end{aligned}.
\end{equation}
Then, the metric tensor of the Bitten Torus can be constructed as:
\begin{equation}
\begin{aligned}
& \left(\mathbf{X}_\theta \cdot \mathbf{X}_\theta\right) d \theta^2+2\left(\mathbf{X}_\theta \cdot \mathbf{X}_\phi\right) d \theta d \phi+\left(\mathbf{X}_\phi \cdot \mathbf{X}_\phi\right) d \phi^2 \\
= & r^2 d \theta^2+(R+r \cos \theta)^2 d \phi^2
\end{aligned}.
\end{equation}
The metric tensor $g$ can also be expressed as:
\begin{equation}
g=\left[\begin{array}{cc}
{r^2} & 0 \\
0 & {(R+r \cos \theta)^2}
\end{array}\right].
\end{equation}
So, we can derive:
\begin{equation}
\begin{aligned}
g^{-1}=\left[\begin{array}{cc}
\frac{1}{r^2} & 0 \\
0 & \frac{1}{(R+r \cos \theta)^2}
\end{array}\right]
\end{aligned},
\end{equation}
\begin{equation}
\begin{aligned}
\quad \frac{\partial g}{\partial \theta}=\left[\begin{array}{cc}
0 & 0 \\
0 & -2(R+r \cos \theta) r \sin \theta
\end{array}\right], \frac{\partial g}{\partial \phi}=\left[\begin{array}{cc}
0 & 0 \\
0 & 0
\end{array}\right]
\end{aligned}.
\end{equation}
We substitute each term into the Geometric Brownian Motion on manifold expressed in equation 1 and obtain the BM on the Bitten Torus:
\begin{equation}
\begin{aligned}
d \theta(t) & =\frac{1}{2}\left(-g^{-1} \frac{\partial g}{\partial \theta} g^{-1}\right)_{11} d t+\frac{1}{4}\left(g^{-1}\right)_{11} \operatorname{tr}\left(g^{-1} \frac{\partial g}{\partial \theta}\right) d t\\
& +\left(g^{-1 / 2}\right)_{11} d B_1(t) \\
 & =-\frac{1}{2} r^{-1} \sin \theta(R+r \cos \theta)^{-1} d t+r^{-1} d B_1(t)
 \end{aligned},
\end{equation}
\begin{equation}
\begin{aligned}
d \phi(t) & =\frac{1}{2}\left(-g^{-1} \frac{\partial g}{\partial \phi} g^{-1}\right)_{22} d t+\frac{1}{4}\left(g^{-1}\right)_{22} \operatorname{tr}\left(g^{-1} \frac{\partial g}{\partial \phi}\right) d t\\
&+ \left(g^{-1 / 2}\right)_{22} d B_2(t) \\
& =0+0+\left|(R+r \cos \theta)^{-1}\right| d B_2(t) \\
& =\left|(R+r \cos \theta)^{-1}\right| d B_2(t)
\end{aligned}.
\end{equation}

$M$'s heat kernel should satisfies the heat equation:
\begin{equation}
\begin{aligned}
\frac{\partial}{\partial t} K_{h e a t}\left(s_{0}, s, t\right)=\frac{1}{2} \Delta_{s} K_{h e a t}\left(s_{0}, s, t\right),\\
\lim _{t \rightarrow 0} K_{\text {heat }}\left(s_{0}, s, 0\right)=\delta\left(s_{0}, s\right), \quad s_{0}, s \in M
\end{aligned},
\end{equation}
where the initial condition is valid in a distributional sense (Berline et al., 2003).\cite{berline2003heat} 
of  is a smooth function $K(x ; y ; t)$ on $M \times M \times R^{+}$.
\onecolumn
\subsection{Application results of Tra-BO and In-BO on Aral sea}
\begin{figure*}[!htb]
	\centering
 \subfigure[True function with initial locations ]{
		\begin{minipage}[b]{0.22\textwidth}
			\includegraphics[width=1\textwidth]{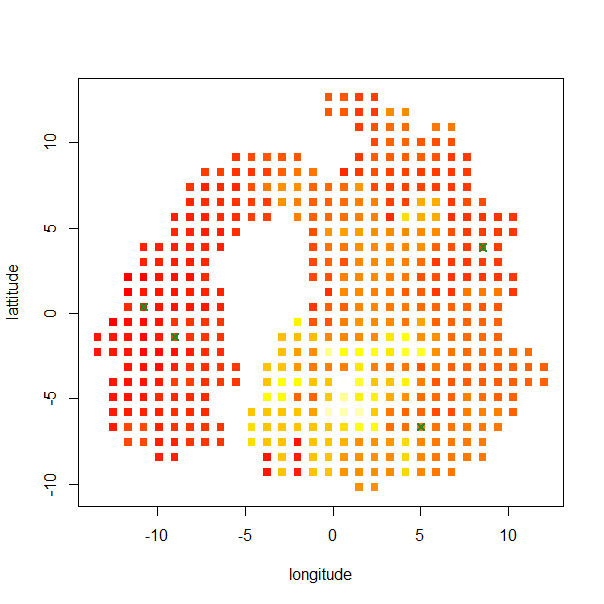} 
		\end{minipage}
		\label{fig:grid_4figs_1cap_4subcap_1}
	}
  \subfigure[Successful application of Tra-BO]{
		\begin{minipage}[b]{0.22\textwidth}
			\includegraphics[width=1\textwidth]{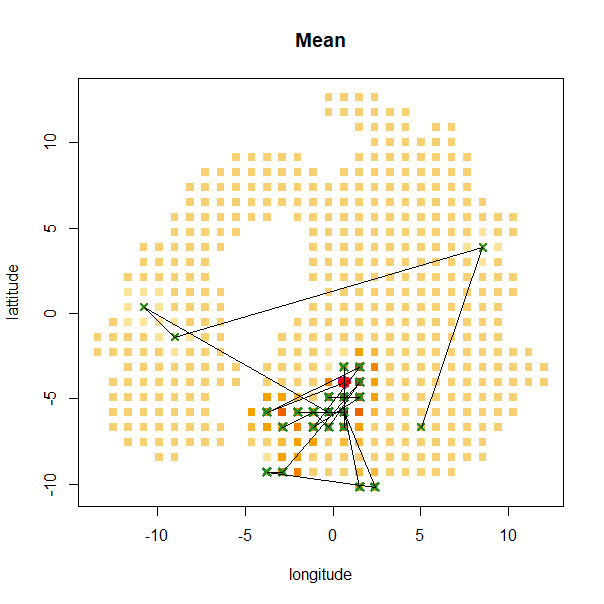} 
		\end{minipage}
		\label{fig:grid_4figs_1cap_4subcap_1}
	}
	\subfigure[In-BO: same initial locations as (b)]{
		\begin{minipage}[b]{0.22\textwidth}
			\includegraphics[width=1\textwidth]{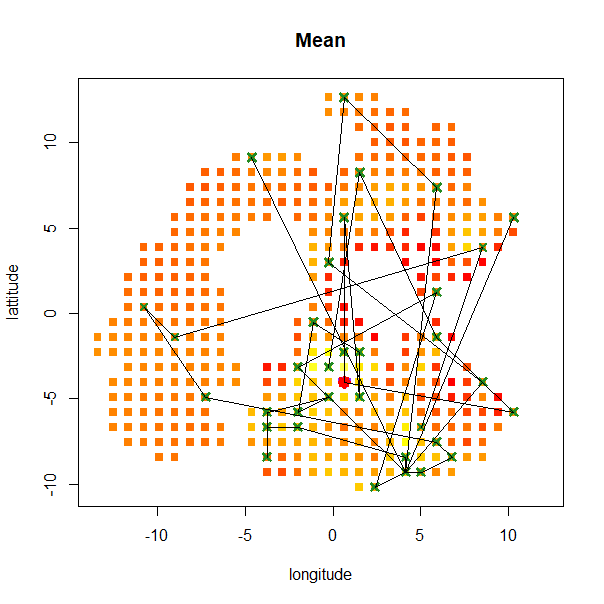} 
		\end{minipage}
		\label{fig:grid_4figs_1cap_4subcap_1}
	}
	\caption{(a): The chlorophyll levels in the Aral sea; The green crosses represent the initial locations for Tra-BO and In-BO shown in (b), (c). (b): Successful application of Tra-BO on Aral sea; the red dot is the global optimum point found by Tra-BO; the green crosses are the continuously updated set of training points. (c): Successful application of In-BO on Aral sea; the red dot is the global optimum point found by In-BO.}
	\label{fig:grid_4figs_1cap_4subcap}
\end{figure*}
\begin{figure*}[!htb]
	\centering
 \subfigure[True function with initial locations]{
		\begin{minipage}[b]{0.22\textwidth}
			\includegraphics[width=1\textwidth]{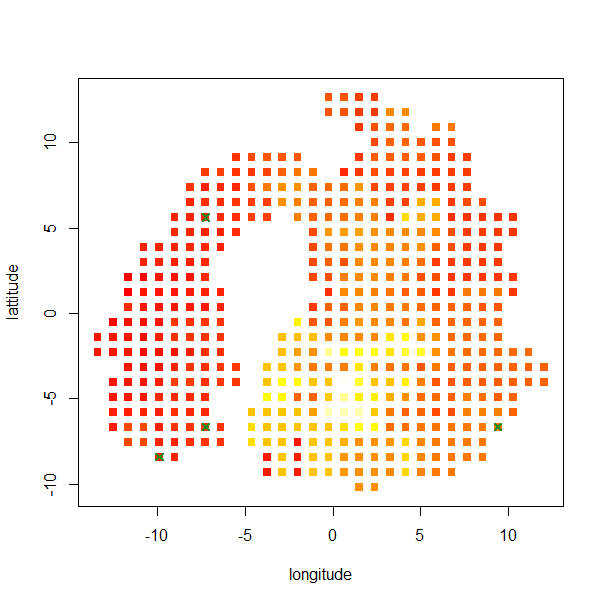} 
		\end{minipage}
		\label{fig:grid_4figs_1cap_4subcap_1}
	}
  \subfigure[Failed application of Tra-BO]{
		\begin{minipage}[b]{0.22\textwidth}
			\includegraphics[width=1\textwidth]{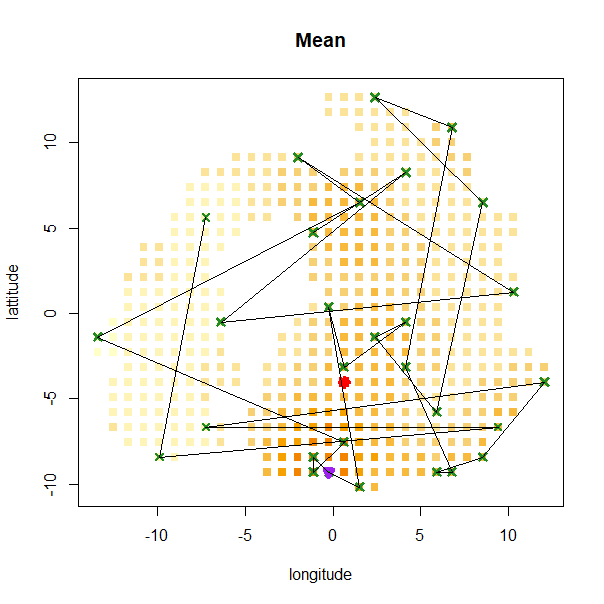} 
		\end{minipage}
		\label{fig:grid_4figs_1cap_4subcap_1}
	}
	\subfigure[In-BO: same initial locations as (b)]{
		\begin{minipage}[b]{0.22\textwidth}
			\includegraphics[width=1\textwidth]{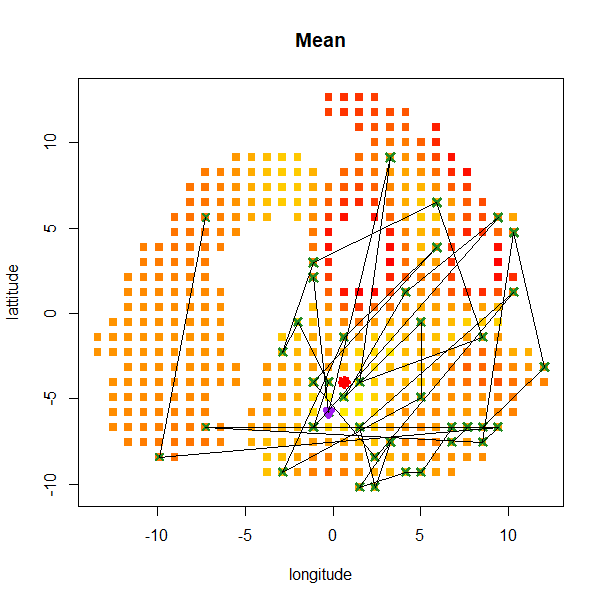} 
		\end{minipage}
		\label{fig:grid_4figs_1cap_4subcap_1}
	}
	\caption{(a): The chlorophyll levels in the Aral sea; The green crosses represent the initial locations for Tra-BO and In-BO shown in (b), (c). (b): Failed application of Tra-BO on Aral sea; the green crosses are the continuously updated set of training points; the purple dot shows the optimum point found by Tra-BO, which is far from the global optimum point shown in the red dot. (c): Application of In-BO on Aral sea with same locations as in (b); the purple dot shows the optimum point found by In-BO, which is around the global optimum point shown in the red dot.}
	\label{fig:grid_4figs_1cap_4subcap}
\end{figure*}

\end{document}